# UNDERSTANDING DEEP LEARNING REQUIRES RETHINKING GENERALIZATION

**Chiyuan Zhang**[*]
Massachusetts Institute of Technology
chiyuan@mit.edu

**Samy Bengio**
Google Brain
bengio@google.com

**Moritz Hardt**
Google Brain
mrtz@google.com

**Benjamin Recht**[†]
University of California, Berkeley
brecht@berkeley.edu

**Oriol Vinyals**
Google DeepMind
vinyals@google.com

## ABSTRACT

Despite their massive size, successful deep artificial neural networks can exhibit a remarkably small difference between training and test performance. Conventional wisdom attributes small generalization error either to properties of the model family, or to the regularization techniques used during training.

Through extensive systematic experiments, we show how these traditional approaches fail to explain why large neural networks generalize well in practice. Specifically, our experiments establish that state-of-the-art convolutional networks for image classification trained with stochastic gradient methods easily fit a random labeling of the training data. This phenomenon is qualitatively unaffected by explicit regularization, and occurs even if we replace the true images by completely unstructured random noise. We corroborate these experimental findings with a theoretical construction showing that simple depth two neural networks already have perfect finite sample expressivity as soon as the number of parameters exceeds the number of data points as it usually does in practice.

We interpret our experimental findings by comparison with traditional models.

## 1 INTRODUCTION

Deep artificial neural networks often have far more trainable model parameters than the number of samples they are trained on. Nonetheless, some of these models exhibit remarkably small *generalization error*, i.e., difference between "training error" and "test error". At the same time, it is certainly easy to come up with natural model architectures that generalize poorly. What is it then that distinguishes neural networks that generalize well from those that don't? A satisfying answer to this question would not only help to make neural networks more interpretable, but it might also lead to more principled and reliable model architecture design.

To answer such a question, statistical learning theory has proposed a number of different complexity measures that are capable of controlling generalization error. These include VC dimension (Vapnik, 1998), Rademacher complexity (Bartlett & Mendelson, 2003), and uniform stability (Mukherjee et al., 2002; Bousquet & Elisseeff, 2002; Poggio et al., 2004). Moreover, when the number of parameters is large, theory suggests that some form of regularization is needed to ensure small generalization error. Regularization may also be implicit as is the case with early stopping.

### 1.1 OUR CONTRIBUTIONS

In this work, we problematize the traditional view of generalization by showing that it is incapable of distinguishing between different neural networks that have radically different generalization performance.

---

[*]Work performed while interning at Google Brain.
[†]Work performed at Google Brain.

**Randomization tests.** At the heart of our methodology is a variant of the well-known randomization test from non-parametric statistics (Edgington & Onghena, 2007). In a first set of experiments, we train several standard architectures on a copy of the data where the true labels were replaced by random labels. Our central finding can be summarized as:

*Deep neural networks easily fit random labels.*

More precisely, when trained on a completely random labeling of the true data, neural networks achieve 0 training error. The test error, of course, is no better than random chance as there is no correlation between the training labels and the test labels. In other words, by randomizing labels alone we can force the generalization error of a model to jump up considerably without changing the model, its size, hyperparameters, or the optimizer. We establish this fact for several different standard architectures trained on the CIFAR10 and ImageNet classification benchmarks. While simple to state, this observation has profound implications from a statistical learning perspective:

1. The effective capacity of neural networks is sufficient for memorizing the entire data set.
2. Even optimization on random labels remains easy. In fact, training time increases only by a small constant factor compared with training on the true labels.
3. Randomizing labels is solely a data transformation, leaving all other properties of the learning problem unchanged.

Extending on this first set of experiments, we also replace the true images by completely random pixels (e.g., Gaussian noise) and observe that convolutional neural networks continue to fit the data with zero training error. This shows that despite their structure, convolutional neural nets can fit random noise. We furthermore vary the amount of randomization, interpolating smoothly between the case of no noise and complete noise. This leads to a range of intermediate learning problems where there remains some level of signal in the labels. We observe a steady deterioration of the generalization error as we increase the noise level. This shows that neural networks are able to capture the remaining signal in the data, while at the same time fit the noisy part using brute-force.

We discuss in further detail below how these observations rule out all of VC-dimension, Rademacher complexity, and uniform stability as possible explanations for the generalization performance of state-of-the-art neural networks.

**The role of explicit regularization.** If the model architecture itself isn't a sufficient regularizer, it remains to see how much explicit regularization helps. We show that explicit forms of regularization, such as weight decay, dropout, and data augmentation, do not adequately explain the generalization error of neural networks. Put differently:

*Explicit regularization may improve generalization performance, but is neither necessary nor by itself sufficient for controlling generalization error.*

In contrast with classical convex empirical risk minimization, where explicit regularization is necessary to rule out trivial solutions, we found that regularization plays a rather different role in deep learning. It appears to be more of a tuning parameter that often helps improve the final test error of a model, but the absence of all regularization does not necessarily imply poor generalization error. As reported by Krizhevsky et al. (2012), $\ell_2$-regularization (weight decay) sometimes even helps optimization, illustrating its poorly understood nature in deep learning.

**Finite sample expressivity.** We complement our empirical observations with a theoretical construction showing that generically large neural networks can express any labeling of the training data. More formally, we exhibit a very simple two-layer ReLU network with $p = 2n + d$ parameters that can express any labeling of any sample of size $n$ in $d$ dimensions. A previous construction due to Livni et al. (2014) achieved a similar result with far more parameters, namely, $O(dn)$. While our depth 2 network inevitably has large width, we can also come up with a depth $k$ network in which each layer has only $O(n/k)$ parameters.

While prior expressivity results focused on what functions neural nets can represent over the entire domain, we focus instead on the expressivity of neural nets with regards to a finite sample. In contrast to existing depth separations (Delalleau & Bengio, 2011; Eldan & Shamir, 2016; Telgarsky, 2016; Cohen & Shashua, 2016) in function space, our result shows that even depth-2 networks of linear size can already represent any labeling of the training data.

**The role of implicit regularization.** While explicit regularizers like dropout and weight-decay may not be essential for generalization, it is certainly the case that not all models that fit the training data well generalize well. Indeed, in neural networks, we almost always choose our model as the output of running stochastic gradient descent. Appealing to linear models, we analyze how SGD acts as an implicit regularizer. For linear models, SGD always converges to a solution with small norm. Hence, the algorithm itself is implicitly regularizing the solution. Indeed, we show on small data sets that even Gaussian kernel methods can generalize well with no regularization. Though this doesn't explain why certain architectures generalize better than other architectures, it does suggest that more investigation is needed to understand exactly what the properties are inherited by models that were trained using SGD.

## 1.2 RELATED WORK

Hardt et al. (2016) give an upper bound on the generalization error of a model trained with stochastic gradient descent in terms of the number of steps gradient descent took. Their analysis goes through the notion of *uniform stability* (Bousquet & Elisseeff, 2002). As we point out in this work, uniform stability of a learning algorithm is independent of the labeling of the training data. Hence, the concept is not strong enough to distinguish between the models trained on the true labels (small generalization error) and models trained on random labels (high generalization error). This also highlights why the analysis of Hardt et al. (2016) for non-convex optimization was rather pessimistic, allowing only a very few passes over the data. Our results show that even empirically training neural networks is not uniformly stable for many passes over the data. Consequently, a weaker stability notion is necessary to make further progress along this direction.

There has been much work on the representational power of neural networks, starting from universal approximation theorems for multi-layer perceptrons (Cybenko, 1989; Mhaskar, 1993; Delalleau & Bengio, 2011; Mhaskar & Poggio, 2016; Eldan & Shamir, 2016; Telgarsky, 2016; Cohen & Shashua, 2016). All of these results are at the *population* level characterizing which mathematical functions certain families of neural networks can express over the entire domain. We instead study the representational power of neural networks for a finite sample of size $n$. This leads to a very simple proof that even $O(n)$-sized two-layer perceptrons have universal finite-sample expressivity.

Bartlett (1998) proved bounds on the fat shattering dimension of multilayer perceptrons with sigmoid activations in terms of the $\ell_1$-norm of the weights at each node. This important result gives a generalization bound for neural nets that is independent of the network size. However, for RELU networks the $\ell_1$-norm is no longer informative. This leads to the question of whether there is a different form of capacity control that bounds generalization error for large neural nets. This question was raised in a thought-provoking work by Neyshabur et al. (2014), who argued through experiments that network size is not the main form of capacity control for neural networks. An analogy to matrix factorization illustrated the importance of implicit regularization.

## 2 EFFECTIVE CAPACITY OF NEURAL NETWORKS

Our goal is to understand the effective model capacity of feed-forward neural networks. Toward this goal, we choose a methodology inspired by non-parametric randomization tests. Specifically, we take a candidate architecture and train it both on the true data and on a copy of the data in which the true labels were replaced by random labels. In the second case, there is no longer any relationship between the instances and the class labels. As a result, learning is impossible. Intuition suggests that this impossibility should manifest itself clearly during training, e.g., by training not converging or slowing down substantially. To our surprise, several properties of the training process for multiple standard achitectures is largely unaffected by this transformation of the labels. This poses a conceptual challenge. Whatever justification we had for expecting a small generalization error to begin with must no longer apply to the case of random labels.

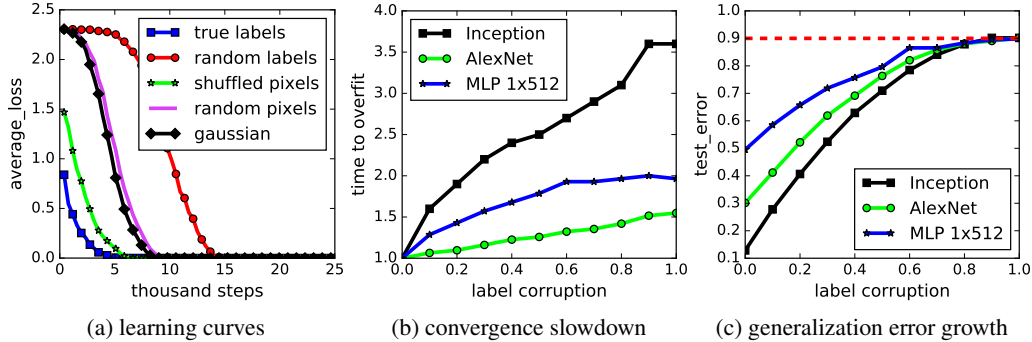

(a) learning curves  (b) convergence slowdown  (c) generalization error growth

Figure 1: Fitting random labels and random pixels on CIFAR10. (a) shows the training loss of various experiment settings decaying with the training steps. (b) shows the relative convergence time with different label corruption ratio. (c) shows the test error (also the generalization error since training error is 0) under different label corruptions.

To gain further insight into this phenomenon, we experiment with different levels of randomization exploring the continuum between no label noise and completely corrupted labels. We also try out different randomizations of the inputs (rather than labels), arriving at the same general conclusion.

The experiments are run on two image classification datasets, the CIFAR10 dataset (Krizhevsky & Hinton, 2009) and the ImageNet (Russakovsky et al., 2015) ILSVRC 2012 dataset. We test the *Inception V3* (Szegedy et al., 2016) architecture on ImageNet and a smaller version of Inception, Alexnet (Krizhevsky et al., 2012), and MLPs on CIFAR10. Please see Section A in the appendix for more details of the experimental setup.

### 2.1 FITTING RANDOM LABELS AND PIXELS

We run our experiments with the following modifications of the labels and input images:

- **True labels**: the original dataset without modification.
- **Partially corrupted labels**: independently with probability $p$, the label of each image is corrupted as a uniform random class.
- **Random labels**: all the labels are replaced with random ones.
- **Shuffled pixels**: a random permutation of the pixels is chosen and then the same permutation is applied to all the images in both training and test set.
- **Random pixels**: a different random permutation is applied to each image independently.
- **Gaussian**: A Gaussian distribution (with matching mean and variance to the original image dataset) is used to generate random pixels for each image.

Surprisingly, stochastic gradient descent with unchanged hyperparameter settings can optimize the weights to fit to random labels perfectly, even though the random labels completely destroy the relationship between images and labels. We further break the structure of the images by shuffling the image pixels, and even completely re-sampling random pixels from a Gaussian distribution. But the networks we tested are still able to fit.

Figure 1a shows the learning curves of the Inception model on the CIFAR10 dataset under various settings. We expect the objective function to take longer to start decreasing on random labels because initially the label assignments for every training sample is uncorrelated. Therefore, large predictions errors are back-propagated to make large gradients for parameter updates. However, since the random labels are fixed and consistent across epochs, the network starts fitting after going through the training set multiple times. We find the following observations for fitting random labels very interesting: a) we do not need to change the learning rate schedule; b) once the fitting starts, it converges quickly; c) it converges to (over)fit the training set perfectly. Also note that "random pixels" and "Gaussian" start converging faster than "random labels". This might be because with random pixels, the inputs are more separated from each other than natural images that originally belong to the same category, therefore, easier to build a network for arbitrary label assignments.

On the CIFAR10 dataset, Alexnet and MLPs all converge to zero loss on the training set. The shaded rows in Table 1 show the exact numbers and experimental setup. We also tested random labels on the ImageNet dataset. As shown in the last three rows of Table 2 in the appendix, although it does not reach the perfect 100% top-1 accuracy, 95.20% accuracy is still very surprising for a million random labels from 1000 categories. Note that we did not do any hyperparameter tuning when switching from the true labels to random labels. It is likely that with some modification of the hyperparameters, perfect accuracy could be achieved on random labels. The network also manages to reach ∼90% top-1 accuracy even with explicit regularizers turned on.

**Partially corrupted labels** We further inspect the behavior of neural network training with a varying level of label corruptions from 0 (no corruption) to 1 (complete random labels) on the CIFAR10 dataset. The networks fit the corrupted training set perfectly for all the cases. Figure 1b shows the slowdown of the convergence time with increasing level of label noises. Figure 1c depicts the test errors after convergence. Since the training errors are always zero, the test errors are the same as generalization errors. As the noise level approaches 1, the generalization errors converge to 90% — the performance of random guessing on CIFAR10.

## 2.2 IMPLICATIONS

In light of our randomization experiments, we discuss how our findings pose a challenge for several traditional approaches for reasoning about generalization.

**Rademacher complexity and VC-dimension.** Rademacher complexity is commonly used and flexible complexity measure of a hypothesis class. The empirical Rademacher complexity of a hypothesis class $\mathcal{H}$ on a dataset $\{x_1, \ldots, x_n\}$ is defined as

$$\hat{\mathfrak{R}}_n(\mathcal{H}) = \mathbb{E}_\sigma \left[ \sup_{h \in \mathcal{H}} \frac{1}{n} \sum_{i=1}^n \sigma_i h(x_i) \right] \quad (1)$$

where $\sigma_1, \ldots, \sigma_n \in \{\pm 1\}$ are i.i.d. uniform random variables. This definition closely resembles our randomization test. Specifically, $\hat{\mathfrak{R}}_n(\mathcal{H})$ measures ability of $\mathcal{H}$ to fit random $\pm 1$ binary label assignments. While we consider multiclass problems, it is straightforward to consider related binary classification problems for which the same experimental observations hold. Since our randomization tests suggest that many neural networks fit the training set with random labels perfectly, we expect that $\hat{\mathfrak{R}}_n(\mathcal{H}) \approx 1$ for the corresponding model class $\mathcal{H}$. This is, of course, a trivial upper bound on the Rademacher complexity that does not lead to useful generalization bounds in realistic settings. A similar reasoning applies to VC-dimension and its continuous analog fat-shattering dimension, unless we further restrict the network. While Bartlett (1998) proves a bound on the fat-shattering dimension in terms of $\ell_1$ norm bounds on the weights of the network, this bound does not apply to the ReLU networks that we consider here. This result was generalized to other norms by Neyshabur et al. (2015), but even these do not seem to explain the generalization behavior that we observe.

**Uniform stability.** Stepping away from complexity measures of the hypothesis class, we can instead consider properties of the algorithm used for training. This is commonly done with some notion of stability, such as *uniform stability* (Bousquet & Elisseeff, 2002). Uniform stability of an algorithm $A$ measures how sensitive the algorithm is to the replacement of a single example. However, it is solely a property of the algorithm, which does not take into account specifics of the data or the distribution of the labels. It is possible to define weaker notions of stability (Mukherjee et al., 2002; Poggio et al., 2004; Shalev-Shwartz et al., 2010). The weakest stability measure is directly equivalent to bounding generalization error and does take the data into account. However, it has been difficult to utilize this weaker stability notion effectively.

## 3 THE ROLE OF REGULARIZATION

Most of our randomization tests are performed with explicit regularization turned off. Regularizers are the standard tool in theory and practice to mitigate overfitting in the regime when there are more parameters than data points (Vapnik, 1998). The basic idea is that although the original hypothesis is too large to generalize well, regularizers help confine learning to a subset of the hypothesis space with manageable complexity. By adding an explicit regularizer, say by penalizing the norm of the optimal solution, the effective Rademacher complexity of the possible solutions is dramatically reduced.

Table 1: The training and test accuracy (in percentage) of various models on the CIFAR10 dataset. Performance with and without data augmentation and weight decay are compared. The results of fitting random labels are also included.

| model | # params | random crop | weight decay | train accuracy | test accuracy |
|---|---|---|---|---|---|
| Inception | 1,649,402 | yes | yes | 100.0 | 89.05 |
| | | yes | no | 100.0 | 89.31 |
| | | no | yes | 100.0 | 86.03 |
| | | no | no | 100.0 | 85.75 |
| (fitting random labels) | | no | no | 100.0 | 9.78 |
| Inception w/o BatchNorm | 1,649,402 | no | yes | 100.0 | 83.00 |
| | | no | no | 100.0 | 82.00 |
| (fitting random labels) | | no | no | 100.0 | 10.12 |
| Alexnet | 1,387,786 | yes | yes | 99.90 | 81.22 |
| | | yes | no | 99.82 | 79.66 |
| | | no | yes | 100.0 | 77.36 |
| | | no | no | 100.0 | 76.07 |
| (fitting random labels) | | no | no | 99.82 | 9.86 |
| MLP 3x512 | 1,735,178 | no | yes | 100.0 | 53.35 |
| | | no | no | 100.0 | 52.39 |
| (fitting random labels) | | no | no | 100.0 | 10.48 |
| MLP 1x512 | 1,209,866 | no | yes | 99.80 | 50.39 |
| | | no | no | 100.0 | 50.51 |
| (fitting random labels) | | no | no | 99.34 | 10.61 |

As we will see, in deep learning, explicit regularization seems to play a rather different role. As the bottom rows of Table 2 in the appendix show, even with dropout and weight decay, InceptionV3 is still able to fit the random training set extremely well if not perfectly. Although not shown explicitly, on CIFAR10, both Inception and MLPs still fit perfectly the random training set with weight decay turned on. However, AlexNet with weight decay turned on fails to converge on random labels. To investigate the role of regularization in deep learning, we explicitly compare behavior of deep nets learning with and without regularizers.

Instead of doing a full survey of all kinds of regularization techniques introduced for deep learning, we simply take several commonly used network architectures, and compare the behavior when turning off the equipped regularizers. The following regularizers are covered:

- **Data augmentation**: augment the training set via domain-specific transformations. For image data, commonly used transformations include random cropping, random perturbation of brightness, saturation, hue and contrast.
- **Weight decay**: equivalent to a $\ell_2$ regularizer on the weights; also equivalent to a hard constrain of the weights to an Euclidean ball, with the radius decided by the amount of weight decay.
- **Dropout** (Srivastava et al., 2014): mask out each element of a layer output randomly with a given dropout probability. Only the Inception V3 for ImageNet uses dropout in our experiments.

Table 1 shows the results of Inception, Alexnet and MLPs on CIFAR10, toggling the use of data augmentation and weight decay. Both regularization techniques help to improve the generalization performance, but even with all of the regularizers turned off, all of the models still generalize very well.

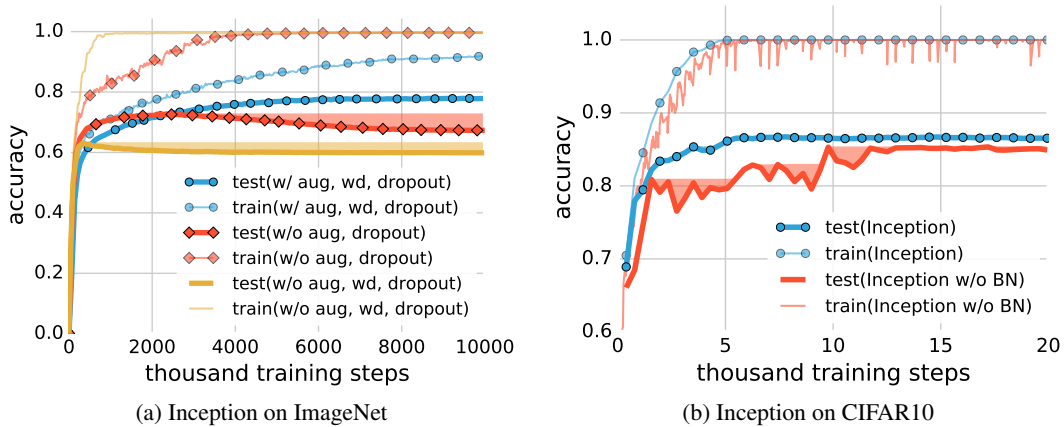

(a) Inception on ImageNet  (b) Inception on CIFAR10

Figure 2: Effects of implicit regularizers on generalization performance. aug is data augmentation, wd is weight decay, BN is batch normalization. The shaded areas are the cumulative best test accuracy, as an indicator of potential performance gain of early stopping. (a) early stopping could potentially improve generalization when other regularizers are absent. (b) early stopping is not necessarily helpful on CIFAR10, but batch normalization stablize the training process and improves generalization.

Table 2 in the appendix shows a similar experiment on the ImageNet dataset. A 18% top-1 accuracy drop is observed when we turn off all the regularizers. Specifically, the top-1 accuracy without regularization is 59.80%, while random guessing only achieves 0.1% top-1 accuracy on ImageNet. More strikingly, with data-augmentation on but other explicit regularizers off, Inception is able to achieve a top-1 accuracy of 72.95%. Indeed, it seems like the ability to augment the data using known symmetries is significantly more powerful than just tuning weight decay or preventing low training error.

Inception achieves 80.38% top-5 accuracy without regularization, while the reported number of the winner of ILSVRC 2012 (Krizhevsky et al., 2012) achieved 83.6%. So while regularization is important, bigger gains can be achieved by simply changing the model architecture. It is difficult to say that the regularizers count as a fundamental phase change in the generalization capability of deep nets.

### 3.1 IMPLICIT REGULARIZATIONS

Early stopping was shown to implicitly regularize on some convex learning problems (Yao et al., 2007; Lin et al., 2016). In Table 2 in the appendix, we show in parentheses the best test accuracy along the training process. It confirms that early stopping could *potentially*[1] improve the generalization performance. Figure 2a shows the training and testing accuracy on ImageNet. The shaded area indicate the accumulative best test accuracy, as a reference of potential performance gain for early stopping. However, on the CIFAR10 dataset, we do not observe any potential benefit of early stopping.

Batch normalization (Ioffe & Szegedy, 2015) is an operator that normalizes the layer responses within each mini-batch. It has been widely adopted in many modern neural network architectures such as Inception (Szegedy et al., 2016) and Residual Networks (He et al., 2016). Although not explicitly designed for regularization, batch normalization is usually found to improve the generalization performance. The Inception architecture uses a lot of batch normalization layers. To test the impact of batch normalization, we create a "Inception w/o BatchNorm" architecture that is exactly the same as Inception in Figure 3, except with all the batch normalization layers removed. Figure 2b compares the learning curves of the two variants of Inception on CIFAR10, with all the explicit regularizers turned off. The normalization operator helps stablize the learning dynamics, but the impact on the generalization performance is only 3∼4%. The exact accuracy is also listed in the section "Inception w/o BatchNorm" of Table 1.

---

[1] We say "potentially" because to make this statement rigorous, we need to have another isolated test set and test the performance there when we choose early stopping point on the first test set (acting like a validation set).

In summary, our observations on both explicit and implicit regularizers are consistently suggesting that regularizers, when properly tuned, could help to improve the generalization performance. However, it is unlikely that the regularizers are the fundamental reason for generalization, as the networks continue to perform well after all the regularizers removed.

## 4 FINITE-SAMPLE EXPRESSIVITY

Much effort has gone into characterizing the expressivity of neural networks, e.g, Cybenko (1989); Mhaskar (1993); Delalleau & Bengio (2011); Mhaskar & Poggio (2016); Eldan & Shamir (2016); Telgarsky (2016); Cohen & Shashua (2016). Almost all of these results are at the "population level" showing what functions of the entire domain can and cannot be represented by certain classes of neural networks with the same number of parameters. For example, it is known that at the population level depth $k$ is generically more powerful than depth $k - 1$.

We argue that what is more relevant in practice is the expressive power of neural networks on a finite sample of size $n$. It is possible to transfer population level results to finite sample results using uniform convergence theorems. However, such uniform convergence bounds would require the sample size to be polynomially large in the dimension of the input and exponential in the depth of the network, posing a clearly unrealistic requirement in practice.

We instead directly analyze the finite-sample expressivity of neural networks, noting that this dramatically simplifies the picture. Specifically, as soon as the number of parameters $p$ of a networks is greater than $n$, even simple two-layer neural networks can represent any function of the input sample. We say that a neural network $C$ can represent any function of a sample of size $n$ in $d$ dimensions if for every sample $S \subseteq \mathbb{R}^d$ with $|S| = n$ and every function $f \colon S \to \mathbb{R}$, there exists a setting of the weights of $C$ such that $C(x) = f(x)$ for every $x \in S$.

**Theorem 1.** *There exists a two-layer neural network with ReLU activations and $2n + d$ weights that can represent any function on a sample of size $n$ in $d$ dimensions.*

The proof is given in Section C in the appendix, where we also discuss how to achieve width $O(n/k)$ with depth $k$. We remark that it's a simple exercise to give bounds on the weights of the coefficient vectors in our construction. Lemma 1 gives a bound on the smallest eigenvalue of the matrix $A$. This can be used to give reasonable bounds on the weight of the solution $w$.

## 5 IMPLICIT REGULARIZATION: AN APPEAL TO LINEAR MODELS

Although deep neural nets remain mysterious for many reasons, we note in this section that it is not necessarily easy to understand the source of generalization for linear models either. Indeed, it is useful to appeal to the simple case of linear models to see if there are parallel insights that can help us better understand neural networks.

Suppose we collect $n$ distinct data points $\{(x_i, y_i)\}$ where $x_i$ are $d$-dimensional feature vectors and $y_i$ are labels. Letting loss denote a nonnegative loss function with $\text{loss}(y, y) = 0$, consider the *empirical risk minimization* (ERM) problem

$$\min_{w \in \mathbb{R}^d} \frac{1}{n} \sum_{i=1}^n \text{loss}(w^T x_i, y_i) \qquad (2)$$

If $d \geq n$, then we can fit any labeling. But is it then possible to generalize with such a rich model class and no explicit regularization?

Let $X$ denote the $n \times d$ data matrix whose $i$-th row is $x_i^T$. If $X$ has rank $n$, then the system of equations $Xw = y$ has an infinite number of solutions regardless of the right hand side. We can find a global minimum in the ERM problem (2) by simply solving this linear system.

But do all global minima generalize equally well? Is there a way to determine when one global minimum will generalize whereas another will not? One popular way to understand quality of minima is the curvature of the loss function at the solution. But in the linear case, the curvature of all optimal solutions is the same (Choromanska et al., 2015). To see this, note that in the case when $y_i$ is a scalar,

$$\nabla^2 \tfrac{1}{n} \sum_{i=1}^n \text{loss}(w^T x_i, y_i) = \tfrac{1}{n} X^T \text{diag}(\beta) X, \qquad \left(\beta_i := \left.\tfrac{\partial^2 \text{loss}(z, y_i)}{\partial z^2}\right|_{z=y_i}, \forall i\right)$$

A similar formula can be found when $y$ is vector valued. In particular, the Hessian is not a function of the choice of $w$. Moreover, the Hessian is degenerate at all global optimal solutions.

If curvature doesn't distinguish global minima, what does? A promising direction is to consider the workhorse algorithm, stochastic gradient descent (SGD), and inspect which solution SGD converges to. Since the SGD update takes the form $w_{t+1} = w_t - \eta_t e_t x_{i_t}$ where $\eta_t$ is the step size and $e_t$ is the prediction error loss. If $w_0 = 0$, we must have that the solution has the form $w = \sum_{i=1}^n \alpha_i x_i$ for some coefficients $\alpha$. Hence, if we run SGD we have that $w = X^T \alpha$ lies in the span of the data points. If we also perfectly interpolate the labels we have $Xw = y$. Enforcing both of these identities, this reduces to the single equation

$$XX^T \alpha = y \qquad (3)$$

which has *a unique solution*. Note that this equation only depends on the dot-products between the data points $x_i$. We have thus derived the "kernel trick" (Schölkopf et al., 2001)—albeit in a roundabout fashion.

We can therefore perfectly fit any set of labels by forming the Gram matrix (aka the *kernel matrix*) on the data $K = XX^T$ and solving the linear system $K\alpha = y$ for $\alpha$. This is an $n \times n$ linear system that can be solved on standard workstations whenever $n$ is less than a hundred thousand, as is the case for small benchmarks like CIFAR10 and MNIST.

Quite surprisingly, fitting the training labels exactly yields excellent performance for convex models. On MNIST with no preprocessing, we are able to achieve a test error of 1.2% by simply solving (3). Note that this is not exactly simple as the kernel matrix requires 30GB to store in memory. Nonetheless, this system can be solved in under 3 minutes in on a commodity workstation with 24 cores and 256 GB of RAM with a conventional LAPACK call. By first applying a Gabor wavelet transform to the data and then solving (3), the error on MNIST drops to 0.6%. Surprisingly, adding regularization does not improve either model's performance!

Similar results follow for CIFAR10. Simply applying a Gaussian kernel on pixels and using no regularization achieves 46% test error. By preprocessing with a random convolutional neural net with 32,000 random filters, this test error drops to 17% error[2]. Adding $\ell_2$ regularization further reduces this number to 15% error. Note that this is without any data augmentation.

Note that this kernel solution has an appealing interpretation in terms of implicit regularization. Simple algebra reveals that it is equivalent to the *minimum $\ell_2$-norm* solution of $Xw = y$. That is, out of all models that exactly fit the data, SGD will often converge to the solution with minimum norm. It is very easy to construct solutions of $Xw = y$ that don't generalize: for example, one could fit a Gaussian kernel to data and place the centers at random points. Another simple example would be to force the data to fit random labels on the test data. In both cases, the norm of the solution is significantly larger than the minimum norm solution.

Unfortunately, this notion of minimum norm is not predictive of generalization performance. For example, returning to the MNIST example, the $\ell_2$-norm of the minimum norm solution with no preprocessing is approximately 220. With wavelet preprocessing, the norm jumps to 390. Yet the test error drops by a factor of 2. So while this minimum-norm intuition may provide some guidance to new algorithm design, it is only a very small piece of the generalization story.

## 6 CONCLUSION

In this work we presented a simple experimental framework for defining and understanding a notion of *effective capacity* of machine learning models. The experiments we conducted emphasize that the effective capacity of several successful neural network architectures is large enough to shatter the training data. Consequently, these models are in principle rich enough to memorize the training data. This situation poses a conceptual challenge to statistical learning theory as traditional measures of model complexity struggle to explain the generalization ability of large artificial neural networks. We argue that we have yet to discover a precise formal measure under which these enormous models are simple. Another insight resulting from our experiments is that optimization continues to be empirically easy even if the resulting model does not generalize. This shows that the reasons for why optimization is empirically easy must be different from the true cause of generalization.

---

[2]This conv-net is the Coates & Ng (2012) net, but with the filters selected at random instead of with k-means.

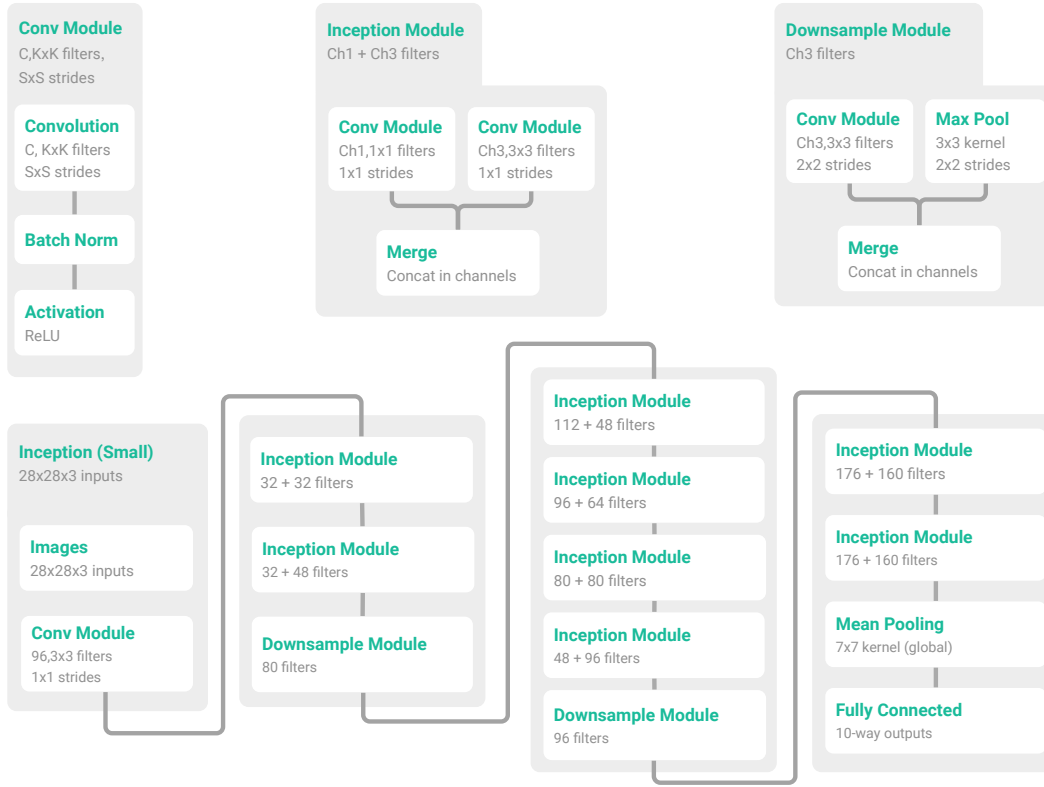

Figure 3: The small Inception model adapted for the CIFAR10 dataset. On the left we show the Conv module, the Inception module and the Downsample module, which are used to construct the Inception architecture on the right.

## A   Experimental setup

We focus on two image classification datasets, the CIFAR10 dataset (Krizhevsky & Hinton, 2009) and the ImageNet (Russakovsky et al., 2015) ILSVRC 2012 dataset.

The CIFAR10 dataset contains 50,000 training and 10,000 validation images, split into 10 classes. Each image is of size 32x32, with 3 color channels. We divide the pixel values by 255 to scale them into $[0, 1]$, crop from the center to get 28x28 inputs, and then normalize them by subtracting the mean and dividing the adjusted standard deviation *independently for each image* with the per_image_whitening function in TENSORFLOW (Abadi et al., 2015).

For the experiment on CIFAR10, we test a simplified Inception (Szegedy et al., 2016) and Alexnet (Krizhevsky et al., 2012) by adapting the architectures to smaller input image sizes. We also test standard multi-layer perceptrons (MLPs) with various number of hidden layers.

The small Inception model uses a combination of 1x1 and 3x3 convolution pathways. The detailed architecture is illustrated in Figure 3. The small Alexnet is constructed by two (convolution 5x5 → max-pool 3x3 → local-response-normalization) modules followed by two fully connected layers with 384 and 192 hidden units, respectively. Finally a 10-way linear layer is used for prediction. The MLPs use fully connected layers. MLP 1x512 means one *hidden layer* with 512 hidden units. All of the architectures use standard rectified linear activation functions (ReLU).

For all experiments on CIFAR10, we train using SGD with a momentum parameter of 0.9. An initial learning rate of 0.1 (for small Inception) or 0.01 (for small Alexnet and MLPs) are used, with a decay factor of 0.95 per training epoch. Unless otherwise specified, for the experiments with randomized labels or pixels, we train the networks without weight decay, dropout, or other forms of explicit regularization. Section 3 discusses the effects of various regularizers on fitting the networks and generalization.

The ImageNet dataset contains 1,281,167 training and 50,000 validation images, split into 1000 classes. Each image is resized to 299x299 with 3 color channels. In the experiment on ImageNet, we use the *Inception V3* (Szegedy et al., 2016) architecture and reuse the data preprocessing and experimental setup from the TENSORFLOW package. The data pipeline is extended to allow disabling of data augmentation and feeding random labels that are consistent across epochs. We run the ImageNet experiment in a distributed asynchronized SGD system with 50 workers.

## B  DETAILED RESULTS ON IMAGENET

Table 2: The top-1 and top-5 accuracy (in percentage) of the Inception v3 model on the ImageNet dataset. We compare the training and test accuracy with various regularization turned on and off, for both true labels and random labels. The original reported top-5 accuracy of the Alexnet on ILSVRC 2012 is also listed for reference. The numbers in parentheses are the best test accuracy during training, as a reference for potential performance gain of early stopping.

| data aug | dropout | weight decay | top-1 train | top-5 train | top-1 test | top-5 test |
|---|---|---|---|---|---|---|
| ImageNet 1000 classes with the original labels | | | | | | |
| yes | yes | yes | 92.18 | 99.21 | 77.84 | 93.92 |
| yes | no | no | 92.33 | 99.17 | 72.95 | 90.43 |
| no | no | yes | 90.60 | 100.0 | 67.18 (72.57) | 86.44 (91.31) |
| no | no | no | 99.53 | 100.0 | 59.80 (63.16) | 80.38 (84.49) |
| Alexnet (Krizhevsky et al., 2012) | | | - | - | - | 83.6 |
| ImageNet 1000 classes with random labels | | | | | | |
| no | yes | yes | 91.18 | 97.95 | 0.09 | 0.49 |
| no | no | yes | 87.81 | 96.15 | 0.12 | 0.50 |
| no | no | no | 95.20 | 99.14 | 0.11 | 0.56 |

Table 2 shows the performance on Imagenet with true labels and random labels, respectively.

## C  PROOF OF THEOREM 1

**Lemma 1.** *For any two interleaving sequences of $n$ real numbers $b_1 < x_1 < b_2 < x_2 \cdots < b_n < x_n$, the $n \times n$ matrix $A = [\max\{x_i - b_j, 0\}]_{ij}$ has full rank. Its smallest eigenvalue is $\min_i x_i - b_i$.*

*Proof.* By its definition, the matrix $A$ is lower triangular, that is, all entries with $i < j$ vanish. A basic linear algebra fact states that a lower-triangular matrix has full rank if and only if all of the entries on the diagional are nonzero. Since, $x_i > b_i$, we have that $\max\{x_i - b_i, 0\} > 0$. Hence, $A$ is invertible. The second claim follows directly from the fact that a lower-triangular matrix has all its eigenvalues on the main diagonal. This in turn follows from the first fact, since $A - \lambda I$ can have lower rank only if $\lambda$ equals one of the diagonal values. □

*Proof of Theorem 1.* For weight vectors $w, b \in \mathbb{R}^n$ and $a \in \mathbb{R}^d$, consider the function $c \colon \mathbb{R}^n \to \mathbb{R}$,

$$c(x) = \sum_{j=1}^n w_j \max\{\langle a, x \rangle - b_j, 0\}$$

It is easy to see that $c$ can be expressed by a depth 2 network with ReLU activations.

Now, fix a sample $S = \{z_1, \ldots, z_n\}$ of size $n$ and a target vector $y \in \mathbb{R}^n$. To prove the theorem, we need to find weights $a, b, w$ so that $y_i = c(z_i)$ for all $i \in \{1, \ldots, n\}$

First, choose $a$ and $b$ such that with $x_i = \langle a, z_i \rangle$ we have the interleaving property $b_1 < x_1 < b_2 < \cdots < b_n < x_n$. This is possible since all $z_i$'s are distinct. Next, consider the set of $n$ equations in the $n$ unknowns $w$,

$$y_i = c(z_i), \quad i \in \{1, \ldots, n\}.$$

We have $c(z_i) = Aw$, where $A = [\max\{x_i - b_i, 0\}]_{ij}$ is the matrix we encountered in Lemma 1. We chose $a$ and $b$ so that the lemma applies and hence $A$ has full rank. We can now solve the linear system $y = Aw$ to find suitable weights $w$. □

While the construction in the previous proof has inevitably high width given that the depth is 2, it is possible to trade width for depth. The construction is as follows. With the notation from the proof and assuming w.l.o.g. that $x_1, \ldots, x_n \in [0, 1]$, partition the interval $[0, 1]$ into $b$ disjoint intervals $I_1, \ldots, I_b$ so that each interval $I_j$ contains $n/b$ points. At layer $j$, apply the construction from the proof to all points in $I_j$. This requires $O(n/b)$ nodes at level $j$. This construction results in a circuit of width $O(n/b)$ and depth $b + 1$ which so far has $b$ outputs (one from each layer). It remains to implement a multiplexer which selects one of the $b$ outputs based on which interval a given input $x$ falls into. This boils down to implementing one (approximate) indicator function $f_j$ for each interval $I_j$ and outputting $\sum_{j=1}^{b} f_j(x)o_j$, where $o_j$ is the output of layer $j$. This results in a single output circuit. Implementing a single indicator function requires constant size and depth with ReLU activations. Hence, the final size of the construction is $O(n)$ and the depth is $b+c$ for some constant $c$. Setting $k = b - c$ gives the next corollary.

**Corollary 1.** *For every $k \geq 2$, there exists neural network with ReLU activations of depth $k$, width $O(n/k)$ and $O(n + d)$ weights that can represent any function on a sample of size $n$ in $d$ dimensions.*

## D  RESULTS OF IMPLICIT REGULARIZATION FOR LINEAR MODELS

Table 3: Generalizing with kernels. The test error associated with solving the kernel equation (3) on small benchmarks. Note that changing the preprocessing can significantly change the resulting test error.

| data set | pre-processing | test error |
|---|---|---|
| MNIST | none | 1.2% |
| MNIST | gabor filters | 0.6% |
| CIFAR10 | none | 46% |
| CIFAR10 | random conv-net | 17% |

Table 3 list the experiment results of linear models described in Section 5.

## E  FITTING RANDOM LABELS WITH EXPLICIT REGULARIZATION

In Section 3, we showed that it is difficult to say that commonly used explicit regularizers count as a fundamental phase change in the generalization capability of deep nets. In this appendix, we add some experiments to investigate how explicit regularizers affect the ability to fit random labels.

Table 4: Results on fitting random labels on the CIFAR10 dataset with weight decay and data augmentation.

| Model | Regularizer | Training Accuracy |
|---|---|---|
| Inception | | 100% |
| Alexnet | Weight decay | Failed to converge |
| MLP 3x512 | | 100% |
| MLP 1x512 | | 99.21% |
| Inception | Random Cropping[1] | 99.93% |
| | Augmentation[2] | 99.28% |

From Table 4, we can see that for weight decay using the default coefficient for each model, except Alexnet, all other models are still able to fit random labels. We also tested random cropping and data augmentation with the Inception architecture. By changing the default weight decay factor from 0.95 to 0.999, and running for more epochs, we observe overfitting to random labels in both cases. It is expected to take longer to converge because data augmentation explodes the training set size (though many samples are not i.i.d. any more).

---

[1]In random cropping and augmentation, a new randomly modified image is used in each epoch, but the (randomly assigned) labels are kept consistent for all the epochs. The "training accuracy" means a slightly different thing here as the training set is different in each epoch. The global average of the online accuracy at each mini-batch on the augmented samples is reported here.

[2]Data augmentation includes random left-right flipping and random rotation up to 25 degrees.